\documentclass[conference]{IEEEtran}
\IEEEoverridecommandlockouts
\usepackage{cite}
\usepackage{amsmath,amssymb,amsfonts}
\usepackage{algorithmic}
\usepackage{graphicx}
\usepackage{textcomp}
\usepackage{soul}
\usepackage{booktabs}
\usepackage{multirow}
\usepackage[table]{xcolor}
\usepackage{adjustbox}
\usepackage{hyperref}
\usepackage[textsize=scriptsize]{todonotes}

\usepackage{xcolor}
\usepackage[normalem]{ulem}

\definecolor{customgreen}{rgb}{0.2, 0.8, 0.2}

\def\BibTeX{{\rm B\kern-.05em{\sc i\kern-.025em b}\kern-.08em
    T\kern-.1667em\lower.7ex\hbox{E}\kern-.125emX}}
\begin{document}

\title{Predicting Outcomes in Video Games with Long Short Term Memory Networks}
\author{
    \IEEEauthorblockN{Kittimate Chulajata\IEEEauthorrefmark{1}, Sean Wu\IEEEauthorrefmark{1}, Fabien Scalzo, Eun Sang Cha}
    \IEEEauthorblockA{
        \textit{Keck Data Science Institute, Pepperdine University, Malibu, California, USA}\\
        Correspondence: \{eunsang.cha\}@pepperdine.edu
    }
    \thanks{\IEEEauthorrefmark{1}Kittimate Chulajata and Sean Wu contributed equally to this research and are co-first authors.}
    \thanks{\IEEEauthorrefmark{2} 
    \href{ https://github.com/kittimateChulajata/Predicting-Outcomes-in-Two-Player-Games-with-LSTM}{Open Source Github Repository}}
}
\maketitle

\begin{figure*}[!t]
    \centering
\includegraphics[width=1.0\textwidth]{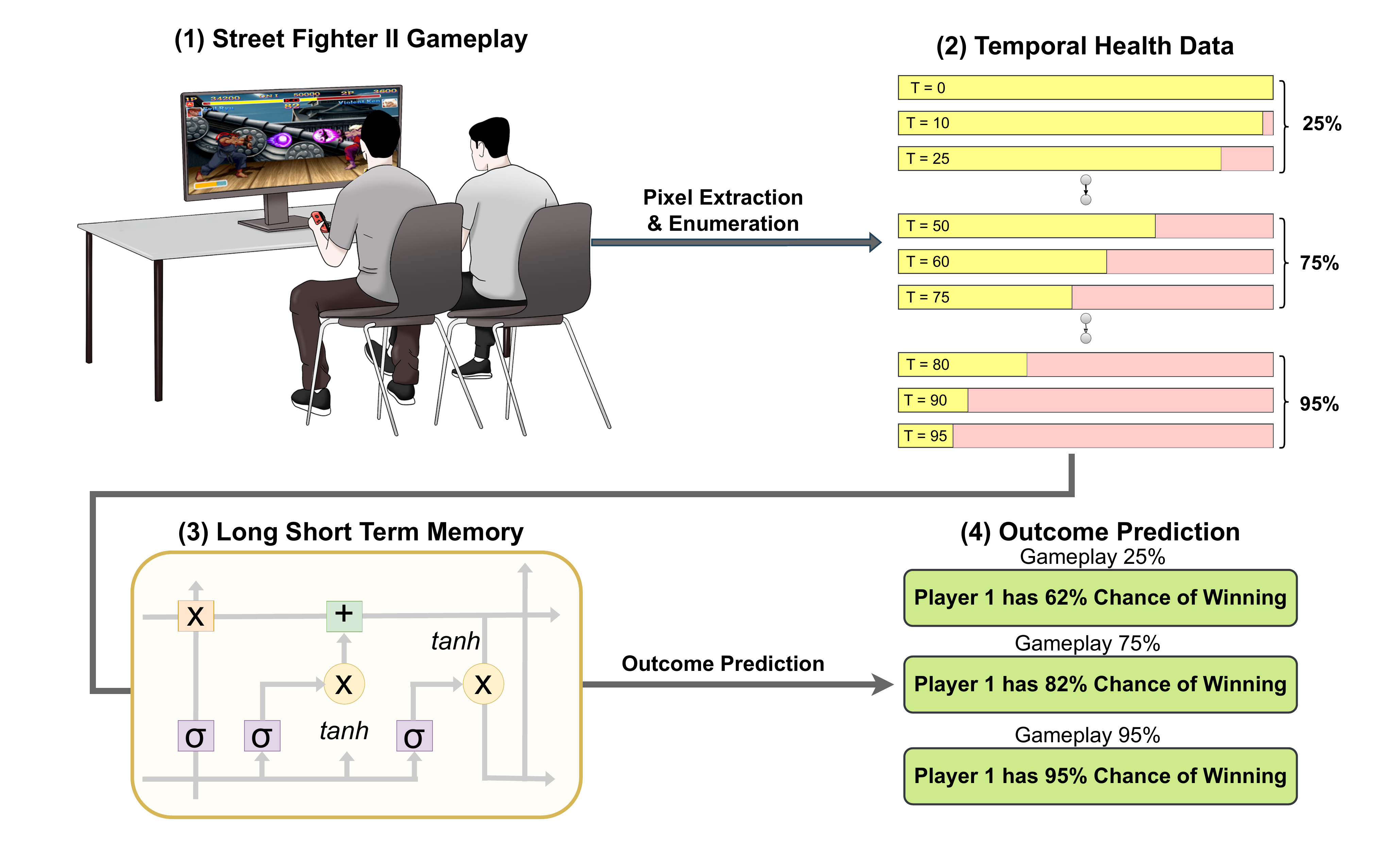}
    End to end visualization of our research in outcome prediction in Super Street Fighter II Turbo with long short term memory networks (LSTMs). Predictions are made from time step 0 to 25\%, 75\%, and 95\% respectively.
    \label{fig:example}
\end{figure*}

\begin{abstract}
 Forecasting winners in E-sports with real-time analytics has the potential to further engage audiences watching major tournament events. However, making such real-time predictions is challenging due to unpredictable variables within the game involving diverse player strategies and decision-making. Our work attempts to enhance audience engagement within video game tournaments by introducing a real-time method of predicting wins. Our Long Short Term Memory Network (LSTMs) based approach enables efficient predictions of win-lose outcomes by only using the health indicator of each player as a time series. As a proof of concept, we evaluate our model's performance within a classic, two-player arcade game, Super Street Fighter II Turbo. We also benchmark our method against state of the art methods for time series forecasting; i.e. Transformer models found in large language models (LLMs). Finally, we open-source our data set and code in hopes of furthering work in predictive analysis for arcade games.

\end{abstract}

\begin{IEEEkeywords}
E-sports, time series analysis, recurrent neural networks, Transformers
\end{IEEEkeywords}

\section{Introduction}

\par Artificial intelligence has become increasingly relevant in recent years, with various applications in forecasting unseen outcomes within sports matches \cite{b23,b29}. Various deep learning approaches have been utilized to forecast the outcomes of competitive matches \cite{b38,b41}. One popular application pertains to audience engagement in electronic sports (E-sports), where an estimated 71 million people watched competitive gaming in 2014 \cite{b39}. In 2015, DOTA 2 offered a 18 million USD prize pool for its annual tournament, considered the largest esport event in history \cite{b40}. The growing field of esports highlights the need for systematic in-game analysis \cite{b42}. These analytics focus on analyzing data from behavioral patterns to identify insightful trends \cite{b43}. Analytics also assists competitors to make an informed decisions and improving understanding of the gameplay tactics. Most importantly, it can simplify in-game dynamics for the viewers, while promoting audience engagement through the use of data driven outcome prediction models. 
\par However, current forecasting of outcomes in sports is limited to pre-match outcome predictions. While win-lose forecasts can directly tie in with the engagement of the audience, traditional sports does not have access to the real-time information necessary to solve this problem. Typically, this gap in real-time analysis and prediction is compensated by live commentary, where experts draw upon their experience to forecast the most probable outcomes of the match. In the traditional sports, similarly to finance, most predictions rely heavily on historical data, such as player statistics or past performances, to predict the outcome \cite{b29}. Thus, these methods lack the ability to take real-time events into consideration. 

\par Esports, on the other hand, is rich in information, opening up the opportunity to solve this real-time prediction problem. Esports offer the advantage of generating real-time data due to ease of access with its digital format. As a consequence, various applications involving video game datasets are emerging. \cite{b30,b18} While there have been multiple attempts to solve this problem, \cite{b31} there have yet to exist a robust yet real-time forecasting method for two player games. 

\par Time series forecasting based on Deep Neural Networks (DNNs) shows remarkable performance both inside  \cite{b32} and outside Esports \cite{b1,b9,b3}. In particular, we focus on the real-time forcasting problem using Long Short-Term Memory (LSTM)\cite{b1}. As proof of concept, we predict the live outcomes of a two-player arcade game, Super Street Fighter II Turbo. Our approach leverages the percentage of health bar loss for each player, allowing for predictions at different stages of the match in real time.

\par Thus, our LSTM-based approach involves real-time, mid-match data sets to predict outcomes at varying time-steps. We experiment at different slices of the game round progression (25\%, 75\%, and 95\%). We attempt to capture the sequential dynamics of two player games as a means of providing greater immersive engagement for the E-sports viewers. By utilizing the percent changes in the health bar of both players, which reflect the extent of damage inflicted, we suggest an approach that is a step towards real-time game play analysis. 

\par Consequently, our contributions are as follows:
\begin{itemize}
    \item \textit{Robust prediction of gaming outcomes in two player games mid-round (25\%, 75\%, \& 95\%) using Long Short Term Memory Networks (LSTMs) as a means of enhancing audience engagement.}
    \item \textit{An open-source data set with code to further research in predictive time series analysis for two-player arcade games. }
\end{itemize}

\section{Existing work}
\par \textit{Quantitative Analysis in E-sports:} Koivisto and Hamari (2019) delve into the significance of gameplay metrics like scores and decision points, illustrating the strong link with player performance and engagement. Their work underscores the intricate relationship between in-game strategies and learning outcomes, emphasizing the importance of these metrics in understanding player behavior and game dynamics \cite{b24}.
\par \textit{Real-Time Esports Prediction:} Hodge (2021) pioneers live analytics, particularly focusing on predicting outcomes in professional matches of games like DOTA 2 \cite{b31} The study employs a diverse range of predictive models, from Logistic Regression to XGBoost, and tests them in major tournaments. Remarkably, their proposed models demonstrate up to 85\% accuracy within the first five minutes of gameplay, marking a significant stride in real-time analytics \cite{b31}.
\par \textit{Comprehensive Match Forecasting:} Yang (2016) aims to enhance prediction accuracy for DOTA 2 matches by incorporating a broad spectrum of pre-match and real-time features. By analyzing a vast set of matches, Yang introduces novel models that boost prediction accuracy, demonstrating the role of real-time data in game progression. This research provides valuable insights for players by combining historical and live data for predictions \cite{b32}.
\par \textit{Neural Learning:} The inception of the Long Short-Term Memory (LSTM) architecture in 1997 marked a significant advancement in time series analysis and deep learning. LSTMs have since been applied to various fields \cite{b1,b2,b3,b4,b5,b6,b7,b8}.
\par \textit{Transformer Attention Models:} Introduced in 2017, the Transformer attention model \cite{b9} has revolutionized natural language processing, replacing LSTMs. Its widespread adoption and recent advancements underscore its effectiveness and versatility across various domains, from audio synthesis to medical data analysis \cite{b9,b10,b11,b12,b13,b14,b15,b16}.
\section{Methodology}
\subsection{Assumptions}
\par We assume repeatability with the gameplay of players. While the gameplay of Super Street Fighter II Turbo uses a variable set of commands for movement, we only utilize the health bar of each player to analyze the progression of the game. We therefore assume that gameplay mechanics, strategies, character uniqueness, movements, player reactions, and decisions are all represented by changes in both players' health at each time step during gaming rounds.

\subsection{Data Collection}
Super Street Fighter II Turbo is a popular two-player fighting game where each player can move, block, and attack to deal damage to the opposing player's health. The goal is to reduce the opponent's health to zero or have more health when the round time expires. The data was collected from 10 Super Street Fighter II Turbo full tournament videos. These were collected from public tournament footage on YouTube. Overall, the data set contains 274,002 rows and 4 columns, where each row corresponds to data extracted from an individual frame, sampled at a rate of 5 frames per second, or a single time-step in a round. This means that a single round may span multiple rows in the data.

\begin{itemize}
  \item \textbf{Winner} - Represents the winner of each round. Used as the target variable for training.
  \item \textbf{Round\_Progression} - Represents the fraction of the round that elapsed at each time step.
  \item \textbf{Player1\_Damaged\%} - Indicates the percentage of health lost by Player 1 due to Player 2's actions at each timestep.
  \item \textbf{Player2\_Damaged\%} - Indicates the percentage of health lost by Player 2 due to Player 1's actions at each timestep.
\end{itemize}

After selecting variables for training, the data was split into rounds. We explain this procedure below. 

\subsection{Splitting Data into Rounds}
\par Our training and prediction models work on a "round basis," meaning that data from each round needed to be separated and extracted from the raw 274,002 rows. Using the column Round\_Progression, we split the data into rounds. For our study, we choose to retain 75\% of the time steps. Other percentages, such as 25\% and 95\% were also tested to gain an understanding of the model's predictive capabilities. With the 75\% time step data, the model was expected to accurately predict the winner of each round before it ended. The percentage of time steps allotted to the model is significant on its performance and real-world effectiveness. For instance, a lower percentage, such as 25\% may not be as accurate, but would be more useful as it could conclude outcomes much quicker. However, a higher percentage, such as 95\%, could provide consistently correct outcomes due to the higher volume of data for the model to interpret, but may not have any useful applications. 
\par In order to split the rows into rounds, we used the column Round\_Progression, where the value 0.0 indicated the beginning of a new round and 100.0 indicated the end. After determining the round "boundaries," we extracted the initial percentage of time steps from each of the rounds. For each round, the winner served as the target variable. Next, the data was split into training and test sets. K-fold cross validation was used with K set to 5 as our data set size was highly constrained. 
 
\par For each iteration, a set of sheets is designated as the test set, and the rest of the data is used for training. Specifically, the test set is determined by selecting data points from sheets indexed from Sheet\_i to Sheet\_(i+3), where i is a variable starting from 2 and incrementing within the loop. The training set consists of data points from sheets not included in the test set. To minimize the risk of data leakage, no overlap between the different sheets or videos was allowed between the sets. After splitting, the train set contained 1154 (roughly 81.2\%) rounds-samples, and the test set contained 267 (roughly 18.7\%) samples.

\par After pre-processing, the class distribution in the data was as follows:
\begin{itemize}
  \item Overall Data: 50.36\% for label 0 \& 49.64\% for label 1.
  \item Training: 53.81\% for label 0 \& 46.19\% for label 1.
  \item Test: 46.82\% for label 0 \& 53.18\% for label 1.
\end{itemize}
\begin{table*}[t]
\centering
\caption{Comparison of LSTM, Transformer, and baseline models in predicting round outcomes.}
\begin{tabular}{lccccc}
\toprule
\multirow{2}{*}{Model} & \multirow{2}{*}{ROC-AUC Scores Across 5 Folds} & \multicolumn{3}{c}{Round Progression} \\
\cmidrule(lr){3-5}
                        &                       & 25\%        & 75\%        & 95\%        \\
\midrule
\multirow{5}{*}{Long Short Term Memory \cite{b1}}   
                        & Fold 1                 & 0.64 ± 0.07 & \textbf{0.94 ± 0.02} & 0.95 ± 0.03 \\
                        & Fold 2                 & \textbf{0.67 ± 0.06} & \textbf{0.94 ± 0.02} & 0.96 ± 0.02 \\
                        & Fold 3                 & 0.65 ± 0.06 & 0.93 ± 0.03 & \textbf{0.98 ± 0.01} \\
                        & Fold 4                 & 0.63 ± 0.05 & 0.93 ± 0.02 & 0.97 ± 0.01 \\
                        & Fold 5                 & 0.65 ± 0.04 & 0.93 ± 0.02 & 0.95 ± 0.02 \\
\midrule
\multirow{5}{*}{Transformer \cite{b9}} 
                        & Fold 1                 & 0.61 ± 0.07 & 0.84 ± 0.05 & 0.87 ± 0.05 \\
                        & Fold 2                 & 0.60 ± 0.06 & 0.92 ± 0.03 & 0.64 ± 0.06 \\
                        & Fold 3                 & 0.59 ± 0.06 & 0.89 ± 0.03 & 0.91 ± 0.03 \\
                        & Fold 4                 & 0.59 ± 0.05 & 0.93 ± 0.02 & 0.83 ± 0.04 \\
                        & Fold 5                 & 0.62 ± 0.05 & 0.89 ± 0.02 & 0.87 ± 0.03 \\
\midrule
\textit{Baseline Models} \\
\midrule
\multirow{5}{*}{Support Vector Machine\cite{svm}}    
                        & Fold 1                 & 0.50 ± 0.07 & 0.51 ± 0.07 & 0.49 ± 0.07 \\
                        & Fold 2                 & 0.52 ± 0.07 & 0.50 ± 0.07 & 0.49 ± 0.06 \\
                        & Fold 3                 & 0.55 ± 0.06 & 0.51 ± 0.06 & 0.49 ± 0.06 \\
                        & Fold 4                 & 0.49 ± 0.05 & 0.52 ± 0.05 & 0.51 ± 0.04 \\
                        & Fold 5                 & 0.48 ± 0.04 & 0.49 ± 0.04 & 0.48 ± 0.04 \\
\midrule
\multirow{5}{*}{Random Forest Classifer \cite{rf}}     
                        & Fold 1                 & 0.49 ± 0.07 & 0.66 ± 0.07 & 0.49 ± 0.07 \\
                        & Fold 2                 & 0.54 ± 0.06 & 0.67 ± 0.06 & 0.49 ± 0.06 \\
                        & Fold 3                 & 0.54 ± 0.06 & 0.63 ± 0.06 & 0.49 ± 0.06 \\
                        & Fold 4                 & 0.53 ± 0.04 & 0.63 ± 0.04 & 0.51 ± 0.04 \\
                        & Fold 5                 & 0.57 ± 0.05 & 0.69 ± 0.04 & 0.48 ± 0.04 \\
\midrule
\multirow{5}{*}{K-Nearest Neighbors \cite{knn}}    
                        & Fold 1                 & 0.49 ± 0.07 & 0.51 ± 0.07 & 0.51 ± 0.07 \\
                        & Fold 2                 & 0.57 ± 0.06 & 0.55 ± 0.06 & 0.54 ± 0.06 \\
                        & Fold 3                 & 0.57 ± 0.06 & 0.55 ± 0.06 & 0.56 ± 0.06 \\
                        & Fold 4                 & 0.48 ± 0.05 & 0.47 ± 0.05 & 0.48 ± 0.05 \\
                        & Fold 5                 & 0.45 ± 0.04 & 0.50 ± 0.04 & 0.49 ± 0.04 \\
\bottomrule
\end{tabular}
\end{table*}
\subsection{Padding Data}
Each data point or round in the splits has the shape (time\_steps, features). time\_steps represents the number of frames for which data has been retained, while features is the number of columns used for training (2 in our case). Due to variation in the temporal sequence of our data, each sample has a varying number of time steps. Thus, the samples needed to be padded so that they all maintained the same number of time steps. We padded all our samples to the same sequence length as a result. When 75\% of the initial time steps were used, the maximum sequence length was 320 time steps. For the Transformer model, we use a padding value of -1 and then generated a Boolean padding mask on it. 
\par Contrary to standard practice in most sequential forecasting applications, we did not use 0 for padding. This is because the feature values for when neither player took damage was already 0. By utilizing the -1 padding, we easily generate masks while retaining legitimate data points. For the LSTM model, the padding value itself was arbitrary and was thus set to 0, because the length of sequences before padding contained our information.

\section{Models}

This section outlines the models that were implemented and tested on the video game data. Results show a high level of performance with LSTMs and Transformer attention-encoders. The implementations of each respective model are detailed below. 

\subsection{Baseline Machine Learning Models}
\subsubsection   {K-Nearest Neighbor Classifier (KNNs)}
\par The K-Nearest Neighbors (KNN) algorithm falls into the category of instance-based learning algorithms. The predictions are made based on the similarity of new data points to existing data points. 

\subsubsection   {Support Vector Classifier (SVMs) }
\par Support Vector Machines (SVMs) finds the optimal hyperplane that most efficiently separates different classes in the feature space. SVM operates by mapping input data into a higher-dimensional feature space using a kernel function. SVM constructs a hyperplane that maximizes the margin in this higher-dimensional space which is distance between the hyperplane and the nearest data points from each class.

\subsubsection   {Random Forest Classifier (RF)}

\par The Random Forest Classifier combines the strength of multiple decision trees to make accurate predictions in classification tasks. It is widely adopted in various domains such as healthcare, finance, and natural language processing due to its robustness and effectiveness.

\subsection{Long Short-Term Memory Classifier}
\par The Long Short-Term Memory (LTSM) model consists of the following hyper parameters:
\begin{itemize}
  \item Output Sequence Length : The output feature dimension of the LSTM layer was set to 8.
  \item Dropout : A dropout of 30\% was utilized to reduce over-fitting.
  \item Classifier Head : The output of our LSTM layer was averaged along the time step dimension. 
\end{itemize}

\subsection{Transformer Classifier}
\begin{figure*}[t]
    \begin{centering} 
    \begin{adjustbox}{center} 
        \includegraphics[width=1.0\textwidth]{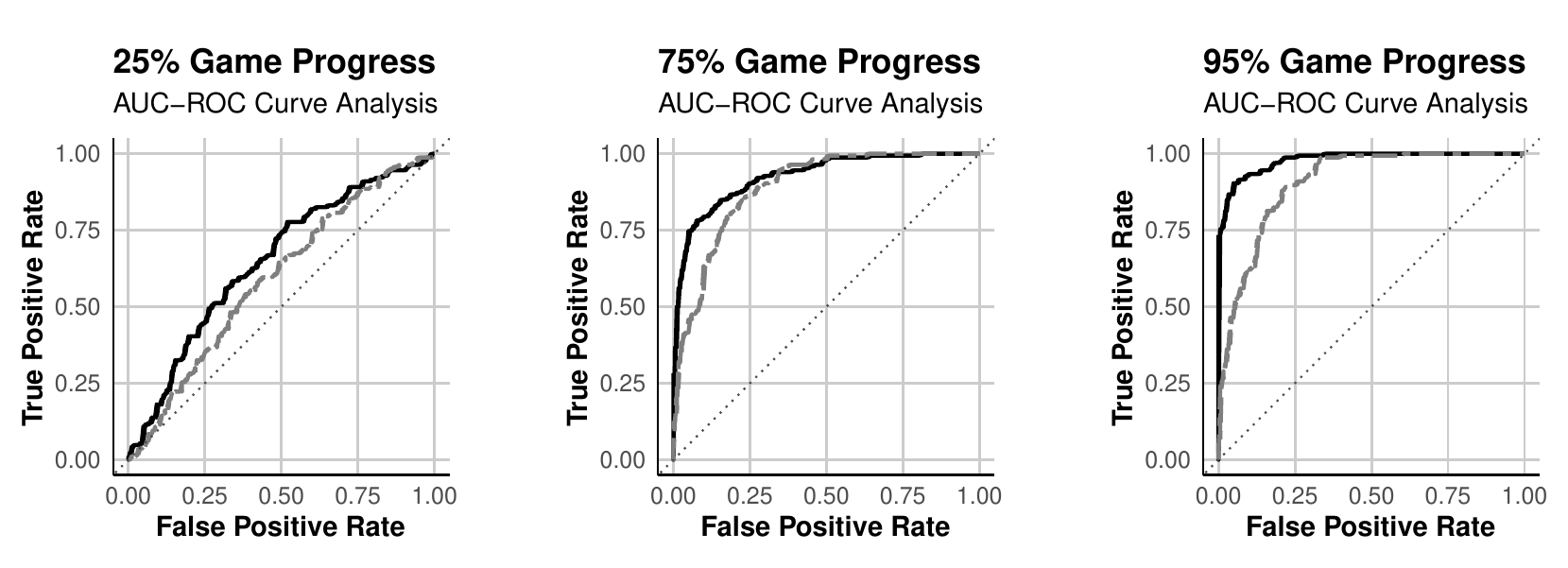} 
    \end{adjustbox}
    \caption{Visualization of ROC Curves comparing LSTM to the Transformer architecture. LSTMs outperform the Transformer to predict round outcomes from 25\%, 75\%, and also 95\%.}
    \label{fig:example}
    \end{centering}
\end{figure*}

\par The Transformer described in \textit{Attention is All You Need} consists of two main branches - an encoder and a decoder. The encoder can access its input sequence in its entirety and provides ”context” that is passed to the decoder \cite{b9} The decoder is an auto regressive component that can only access values it has generated previously, as well as the encoder output. The decoder predicts the next value in the sequence, using the ”context” provided by the encoder and its own outputs from earlier time steps, which is similar to how LSTMs work. The Transformer model in our benchmark tests utilizes only the Transformer encoder to process time steps from the rounds, inspired by encoder-only models like BERT \cite{b20} or RoBERTa \cite{b21} that learn to predict tokens that have been masked out of input sequences. For our use case, the contextual information from the encoder is averaged along the time step dimension, like in the LSTM model, and passed to the classifier head for final prediction. However, it should be noted that models like BERT use an alternative approach where a special token called [CLS] is pre-appended to input sequences \cite{b20} The state of this token is extracted from the encoder output to serve as the final classification result produced by the model. The hyperparameters of our Transformer encoder-based classifier is as follows:
\begin{itemize}
  \item Linear Embedding Layer: The output dimension of this model was set to 8.
  \item Positional Encoding: Dropout of 30\% and maximum vocabulary size of 722 are used in the positional encoding.
  \item  Encoder: The dimension of the feed forward network at the end of the encoder was set to 8, the number of heads in our multi-head attention was set to a size of 4.
\end{itemize}

\subsection{Training Procedure}
\par For both classifiers, the Adam optimization algorithm \cite{b22} was used. The binary cross-entropy function with logits was chosen as the loss function to minimize. The binary cross-entropy loss function is given by the following equation:

$$ -\frac{1}{N}\sum_{i = 1}^{N}{y_{i}\log{\hat{y_{i}}+ (1-y_{i})\log{(1-y_{i})}}} $$
The learning rate for the Transformer model was set to 0.0006, while the LSTM learning rate was set to 0.001. For both models, an L2 regularization factor of 1e-4 was imposed on the optimizer. The batch size for training was set to 28 for the Transformer model and 64 for the LSTM model. Both models were trained for 500 epochs across 5 folds.

\section{Results}

\par In this study, the output logits were converted to probabilities using the sigmoid activation function, allowing us to properly benchmark the models. For the training sets, metrics were collected and averaged for each epoch over 5 training runs. For the test set, the model was evaluated once at the end of each training run - results were later averaged as well.

\par Within 400 epochs of training, neither model converged. However, judging by loss alone, the Transformer model performed better on the training set relative to its test set compared to the LSTM classifier. The LSTM model achieved lower test loss than the Transformer model due to the latter over-fitting. 

\begin{table}[h]
\centering
\caption{LSTM and Transformer Inference Time Across Five Folds}
\label{tab:ModelPerformanceEnhanced}
\begin{tabular}{lccc}
\toprule
& \multicolumn{3}{c}{Round Progression (\%)} \\
\cmidrule{2-4}
Model & 25\% & 50\% & 95\% \\
\midrule
LSTM (ms)& \(0.781 \pm 0.241\) & \(2.203 \pm 0.752\) & \(2.768 \pm 0.945\) \\
Transformer (ms)& \(1.737 \pm 3.326\) & \(2.058 \pm 3.519\) & \(2.428 \pm 4.087\) \\
\bottomrule
\end{tabular}
\end{table}

\section{Discussion}
\par Both Long Short-Term Memory Networks (LSTMs) and Transformer Attention Models were shown to be effective in real-time forecasting. The LSTM classifier exhibits marginal performance gains in key indicators such as AUC (Area Under the Curve). In an effort to test the models' robustness, we also evaluated their performance on a round progression of 95\%. Interestingly, this evaluation did not lead to a significant improvement in performance. However, we give metrics for models trained on just 25\% percent of the initial time steps in the rounds. We can see that performance significantly degrades with a lower percentage of data points included. With regards to our method, prior research shows (Sak et al. 2014) that LSTMs being sequential, have difficulty processing long sequences \cite{b19}. While LSTMs were proficient for the smaller-scale applications, its performance could be largely improved by further exploration into its architecture. The Transformer encoder can effectively handle parallelization and sequence length issues. This was due to its distinctive components, such as positional encoding and diagonal look-ahead masks, allowing it to handle large sequential tasks.  

\subsection{Limitations}
While we acknowledge the small size of our data set as a key limitation in our study, as a potential solution, we refer to Shorten and Khoshgoftaar (2019), \cite{b34} which demonstrate the potential of data augmentation techniques tailored to specific data set characteristics. Explorations of these techniques that introduce perturbations in player actions might enhance each model's ability to generalize within diverse gameplay scenarios. Furthermore, data from Youtube videos typically highlight key moments in a game round. Thus, the gameplay behaviors may not be representative of all Super Street Fighter II Turbo gameplay as a whole. We acknowledge that our data set may be limited to highlights or gameplay between professionals in unusual moments. These are all outliers, and therefore it is unclear whether our model would be effective on determining outcomes between casual players. As a solution, future work will involve collecting data from a broader range of participants, ranging from new entrants to experts. 

\section*{Conclusion}
\par In conclusion, we demonstrate the predictive capacity of LSTMs for player outcomes in two player games. Our preliminary work can further engage the audience in anticipation of gaming outcomes. Audience-centric considerations are necessary in entertainment; predictions on gameplay bridge the gap between competition and the viewer's immersion in entertainment experience. Future work may involve in-game pose estimation to further predictive performance. This may also be useful in predicting behavioral or emotional changes for the players. For example, one could apply our predictive analysis method towards the detection of 'skill drop moments' during game play for an understanding of human computer interactions (HCI). We also suggest hybrid architectures which combine the strength of different models for future work. Research by Pham et al. (2018) \cite{b35} delves deeper into or ensemble learning, suggesting that hybrid models can ultimately outperform individual architectures \cite{b34}. We hope our work serves as a groundwork for furthering research in predicting outcomes in other games within E-ports.

\section*{Acknowledgments}

\par We would like to acknowledge Capcom Co., Ltd for gameplay videos of Super Street Fighter II Turbo. In compliance to Capcom Co., Ltd’s policy regarding usage of gaming content, our open-source data set is solely intended for the purposes of research and education. We would also like to thank the Keck Foundation Grant in support of this research.

\end{document}